# Evolutionary Deep Reinforcement Learning Using Elite Buffer: A Novel Approach Towards DRL Combined with EA in Continuous Control Tasks

Marzieh Sadat Esmaeeli

m.s.esmaeeli1995@gmail.com

Hamed Malek

h_malek@sbu.ac.ir

*Faculty of Computer Science and Engineering, Shahid Beheshti University, Tehran, Iran*

**Abstract**

Despite the numerous applications and success of deep reinforcement learning in many control tasks, it still suffers from many crucial problems and limitations, including temporal credit assignment with sparse reward, absence of effective exploration, and a brittle convergence that is extremely sensitive to the hyperparameters of the problem. The problems of deep reinforcement learning in continuous control, along with the success of evolutionary algorithms in facing some of these problems, have emerged the idea of evolutionary reinforcement learning, which attracted many controversies. Despite successful results in a few studies in this field, a proper and fitting solution to these problems and their limitations is yet to be presented. The present study aims to study the efficiency of combining the two fields of deep reinforcement learning and evolutionary computations further and take a step towards improving methods and the existing challenges. The "Evolutionary Deep Reinforcement Learning Using Elite Buffer" algorithm introduced a novel mechanism through inspiration from interactive learning capability and hypothetical outcomes in the human brain. In this method, the utilization of the elite buffer (which is inspired by learning based on experience generalization in the human mind), along with the existence of crossover and mutation operators, and interactive learning in successive generations, have improved efficiency, convergence, and proper advancement in the field of continuous control. According to the results of experiments, the proposed method surpasses other well-known methods in environments with high complexity and dimension and is superior in resolving the mentioned problems and limitations.

## 1. Introduction

Despite the numerous applications and success of deep reinforcement learning in many control tasks, it still suffers from many problems, including temporal credit assignment with sparse reward, absence of effective exploration, and a brittle convergence that is extremely sensitive to the hyperparameters of the problem. As the complexity and dimension of the problem's data and environment increase, these limitations become more effective in preventing the desired outcome and the necessary functional efficiency. In general, reinforcement learning and deep reinforcement learning fields still face many deficiencies and limitations in complex problems, including real-world scenarios. However, researchers aim to obtain reinforcement learning methods that possess accuracies and efficiencies similar to human beings and, ultimately, surpass human efficiency by avoiding human biasing and error [1-2].

Reinforcement learning in the human brain (as the primary theoretical framework to determine values and generalize experiences), possesses superior capabilities, including assigning temporal credit and tracing eligibility, discerning the received reward from sequential tasks (without stage-wise reward), learning in sparse reward environments, learning in strategic environments while interacting, calculating hypothetical outcomes for unfulfilled tasks and updating their respective value functions, simulating future tasks based on hypothetical experiences, swiftly and optimally tuning the learning and discount rate, and learning based on combining model-based and model-free reinforcement learning. These capabilities are among the noteworthy notions that currently face significant challenges and limitations in the field of deep reinforcement learning, and the relative literature has yet to present a proper solution for implementing an efficient mechanism for them [3-9].

On the other hand, some researchers believe in the use of evolutionary computation methods to solve the three fundamental limitations of deep reinforcement learning (i.e., temporal credit assignment with sparse reward, absence of effective exploration, and a hyperparameter-sensitive brittle convergence). Although evolutionary computations have superior performance in dealing with these limitations, in highly-complex samples and problems that require the optimization of a large number of parameters, they require a high amount of data and still face deficiencies and limitations.

Few studies in the literature have investigated the combination of reinforcement learning and evolutionary computations (e.g., the ERL method [1]) and the combination of deep reinforcement learning and evolutionary computations (e.g., the CEM-RL method [10]). The main concept of these methods was the simultaneous use of effective exploration capacity with diversified policy sets, and the consistency of population-based approaches in evolutionary computations, along with the capacity of the gradient in the efficiency of higher-tier samples and faster learning in Off-Policy reinforcement learning. While the latest approaches presented in this field were capable of improving conditions and yielding success in a limited number of fields, they were not recognized as ideal solutions. In addition, the combination of these two fields has given rise to many controversies that offer many reasons for their objection.

Evolutionary Deep Reinforcement Learning (EDRL) is among the latest emerging fields of study in machine learning and optimization. The earliest studies in this field were initiated in 2018. However, EDRLs have many supporters and opposers, and as soon as they were presented, many objections pointed out their inefficiency due to the high number of hyperparameters, high temporal cost, and high execution resources costs. Nevertheless, this field was still studied owing to the high capacity of deep reinforcement learning and evolutionary computations in solving complex problems and resolving each other's deficiencies. On the other hand, due to advancements in hardware and execution resources such as GPUs and cloud-based services, the infrastructural costs are no longer a cause for barricading or decelerating advancements in this field as in previous decades.

This field can be considered a 'population-driven guide' and a Lamarckian mechanism in terms of reinforcement learning and evolutionary computation, respectively [1]. In other words, reinforcement learning methods learn from an actor's lifetime (i.e., the actor's experiences while executing a task), while evolutionary methods learn from the death of an actor (the calculated fitness function at the end of a complete execution). As discussed in the study that proposed the ERL method, the primary mechanism in this field is "*direct learning from the high resolution of individual experiences corresponding to the increase in the maximum long-term return using low-resolution eligibility criteria*" [1].

In light of the existence of a rich background in computational neuroscience and the capabilities of the human mind (which has consistently inspired and assisted artificial intelligence in diverse and extensive aspects), the present study has gained inspiration from a number of capabilities and features of the human brain in reinforcement learning to improve the methods for deep reinforcement learning. The main objective of the paper is to present a method based on the combination of deep reinforcement learning and evolutionary computations to 1) study and evaluate the efficiency of combining these two fields and 2) improve the existing methods. The algorithm proposed in this paper, with the condition of the existence of a neural basis for interactive learning and the conformity of this mechanism with evolutionary computations, attempts to improve the total performance return of the method in continuous control environments with high dimensions and complexities. Moreover, this algorithm attempts to compensate for problems such as temporal credit assignment, dependency on initial parameters, and brittle convergence using ideas from interactive learning and learning from hypothetical outcomes.

Since the present study combines multiple research fields, it can be assessed according to several aspects. In terms of reinforcement learning, the present study possesses an actor-critic approach that utilizes value functions in conjunction with policy search [11]. Moreover, in terms of deep reinforcement learning, it is considered among the model-free off-policy methods called "Patient Deep Reinforcement Learning" methods [12]. Based on the classification provided for evolutionary computations, the proposed method will be considered in the design perspective [13].

In terms of neuroscience, the present study proposes the idea of the elite buffer based on investigating mechanisms of the human brain in the field of learning, including interactive learning, learning from experiences and hypothetical outcomes, analysis in rest time (i.e., the capability to simulate the outcome for potential actions that can be chosen), temporal credit assignment, and eligibility trace. To learn more about the mentioned mechanisms, you may refer to [4], [8-9], [14], and [15-17].

The remainder of the paper is as follows: Section 2 includes a brief explanation of previous works in this field. In Section 3, we first describe the efficiency of the field and the origin of the proposed idea and then describe the methodology in detail. Some descriptions of simulators, experiments, and obtained results are provided in Section 4. Further discussion regarding the efficiency and the assessment of the proposed method is presented in Section 5. Finally, the research is concluded in Section 6.

## 2. Related Works

In general, the mentioned algorithms in deep reinforcement learning were inspired by the Deep Q Network (DQN) method and indicated an evolutionary flow based on the experiences of their predecessors. The DDPG algorithm was an advanced and improved version of DQN based on the actor-critic concept. While DQN achieved near human-level accuracy in Atari games, it only yielded good performance in problems with high-dimensional observation space and low-dimensional discrete action space [18].

The TD3 method attempted to increase the stability and performance of DDPG by considering the function approximation error. In [19], Scott Fujimoto et al. proved that "*the value estimate in deterministic policy gradients under some basic assumptions will be an overestimation*".

Evolutionary Reinforcement Learning (ERL) is a method proposed based on the combination of evolutionary computations and the gradient descent method. This method attempted to improve the conditions of temporal credit assignment and provide better learning in sparse reward problems [1]. Moreover, the CERL method, which was the continuation of research and activities of the presenters of ERL, employed similar concepts towards improving the process of exploration and exploitation in the environment while learning [20].

The Cross-Entropy Method-Reinforcement Learning (CEM-RL) algorithm was among the widely successful methods in this field. The main idea of this method was to combine the Cross-Entropy Method

(CEM) with one of the deep reinforcement learning methods of DDPG or TD3. Therefore, it was also introduced as CEM-DDPG and CEM-TD3 [10].

In the multi-task trained method, the focus was on providing a comprehensive model that offered training in multiple environments (that can be successfully executed in unobserved environments) and avoided catastrophic forgetting in the training of the neural networks. Using evolutionary algorithms and proper exploration, [21] successfully solved the mentioned problem, avoided falling into local optima, and eventually presented a multi-task trained model.

Despite the partial success of these methods in improving this field of research, they still face significant problems and limitations. Specifically, DDPG suffered from relative instability due to overestimation bias caused by updating the critic network and high sensitivity to parameter tuning. On the other hand, while overestimation was improved to a large extent in TD3, it still suffered from the problems in deep reinforcement learning, which caused the movement towards combining with evolutionary computation methods. Moreover, even though CEM-RL succeeded in decreasing the variance in the training process, the ideas of the importance mixing and adding noise were not successful. In addition, while the multi-task trained method combined deep reinforcement learning and evolutionary algorithm to improve conditions, provide a better exploration of the environment, and avoid dependency on the initial point of the model, it did not present a suitable achievement. Finally, despite the fact that the CERL algorithm succeeded in obtaining proper and fitting results, it had high sensitivity to hyperparameters. Additionally, as the capability of this method was dependent on obtaining ideal parameters, it required significant computation resources and was inconveniently time-consuming [1][22][18-21].

## 3. Methodology

In this section, we first describe the concept and the mechanism for the proposed buffer in separate subsections. Then, we present the proposed method of the study in detail.

### 3.1. The Proposed Elite Buffer

Our familiarization with the existing neural mechanisms has yielded the emergence of the idea of creating an auxiliary buffer for an EDRL method in the objective environments of the field of our research.

The interactive learning process has increased confidence in the use of evolutionary processes with the population members in the form of DRL. The auxiliary buffer, which is entitled "elite buffer" in the present study, includes a number of the most different sequences of successful experiences of the population members. In other words, this buffer includes a number of sequences of complete experiences (from the initial point to the end of one complete iteration of execution in the environment) selected from the most successful experiences of all the members of the population. In the gaps between generations (which reminiscents the rest time in the learning process of the human brain), this buffer is provided to the elite member of the population as a hypothetical experience. Since the elite buffer includes the experiences collected from all the population members, it is taken as a hypothetical experience sequence for the elite member. The elite member can be trained using this buffer, and the results obtained from this training yield a preparation and improvement in the adopted policy in similar events. Moreover, as the experiences in the buffer are complete sequences of experiences selected from executions with high final rewards, the elite member can exploit the advantages of these actions in the presence of sequential events with proper rewards or proper actions with high delay (i.e., the temporal credit assignment problem). In fact, in the case of the existence of such desired events, they will be parts of a very successful execution. In this case, the use of this buffer is a way to utilize and prioritize sequential events with proper rewards, proper events with delayed rewards, and single-action events with proper rewards. On the other hand, due to selecting the sequences in the buffer from the experiences of all members of the population, biasing caused by correlation in the data will be eliminated.

In addition, the use of the evolutionary process and executing crossover and mutation operators in conjunction with the elite buffer yields the necessary diversity for exploration in the environment and the adopted policy.

### 3.2. The Proposed Method

The method proposed in this study is an EDRL algorithm with an emphasis on the idea of using the elite buffer, which is inspired by the concepts of simulating the hypothetical outcomes and human interactive learning. The general flow of this algorithm is as follows: first, one of the authorized DRL models is initialized and created for all the members of the population with similar architecture and different random parameters. Each member of the population has a distinct replay buffer. Then, each member is executed in the environment with a specific number of iterations and is trained based on the selected DRL model, and their experience sequence is stored in their respective replay buffer. Next, each member is evaluated, and their respective fitness function is computed. Afterward, the elite buffer is created using the replay buffer of the members, and the best member (based on their fitness functions) is selected as the elite member. A copy of the elite member is directly placed in the next generation, and another copy is transferred to the next generation after being trained using the elite buffer. From the *n* members of the current generation, *n-2* members are selected as parents to apply the crossover operator and then the mutation operator. Following the application of operators, offsprings join the next generation. This evolutionary process cycle is repeated until the termination condition is satisfied. Finally, the elite member of the last execution generation is introduced as the solution to the problem. The general diagram of the proposed algorithm entitled "Evolutionary Deep Reinforcement Learning using Elite Buffer" (EDRL-EB) is shown in Figure 1.

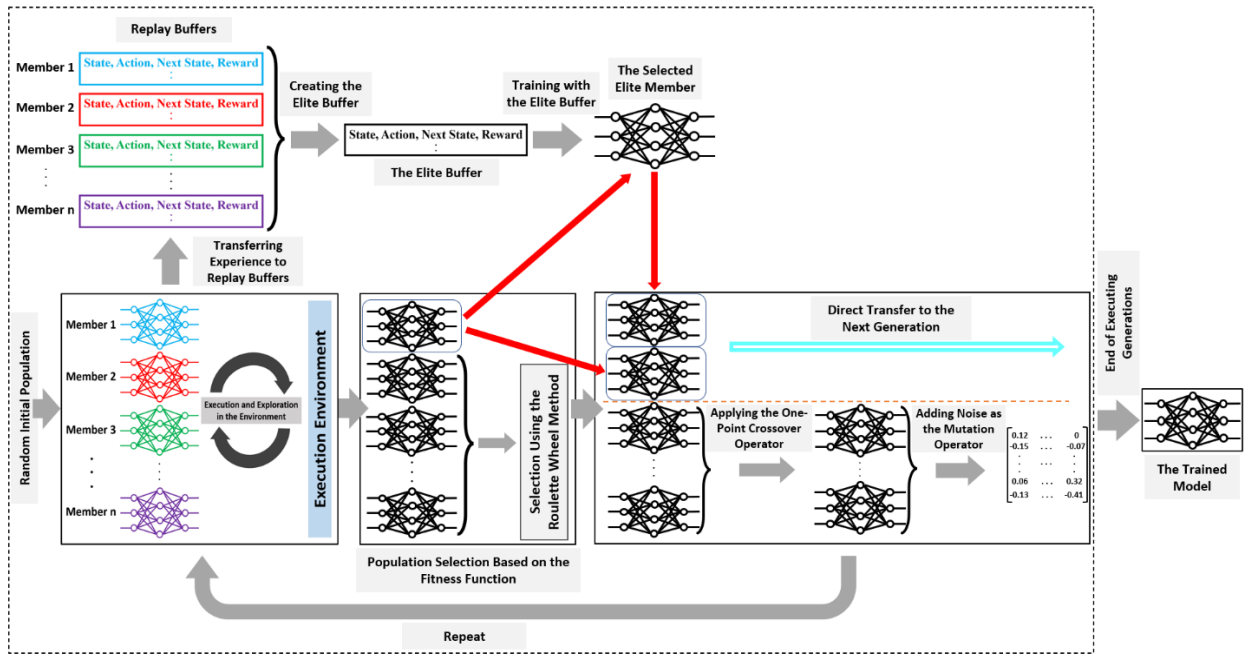

Figure 1: The general diagram of the EDRL-EB algorithm. It shows the evolutionary process including DRL models as population's members and utilization of the elite buffer. Process display style inspired by [21]

### 3.2.1. The Evolutionary Process

The evolutionary algorithm employed in this method is a simple genetic algorithm. In the utilized evolutionary process, the population members of the first generation are initialized using random parameters. After each member is executed for a specific number of iterations, it is evaluated to determine its fitness function. For the determination of the fitness function, each member is executed in the environment a specific number of times, and the mean value for their total obtained rewards in mentioned executions is selected as the fitness function of that member. Then, the best member is selected as the elite member, and a copy of it is identically placed in the next generation. In addition, another copy is first trained using the elite buffer (which is created using the replay buffers of the members of the current generation) and is then placed in the next generation. Next, using the roulette wheel selection method, *n-2* members are selected as parents from the members of the current generation to create the members of the next generation. Afterward, the crossover operator is applied to parents in a pairwise manner, followed by the mutation operator on the born members. The new population will then replace the members of the previous generation. The condition for the termination of the evolutionary process is the execution of the specified number of generations, which is determined by calculating the overall number of interactions of the members with the environment in relation to the number of members of the population. This procedure is repeated until the termination condition is obtained. In the end, the elite member of the final execution generation is introduced as the solution to the problem.

### 3.2.2. Members of the Population

The members of the population in our proposed method are DRL models. In this algorithm, it is possible to use both DDPG and TD3 deep reinforcement learning methods. Each member of the population is a DRL model that is initialized using distinct random values at the beginning of the first generation.

### 3.2.3. The Elite Buffer

Before discussing the mechanism of the creation of the elite buffer, the placement of each transition quadruple $(s_t, a_t, s_{t+1}, r_t)$ in each row of the buffer is shown in Figure 2 to clarify the mechanism. In the following, the mechanism to create the elite buffer will be discussed.

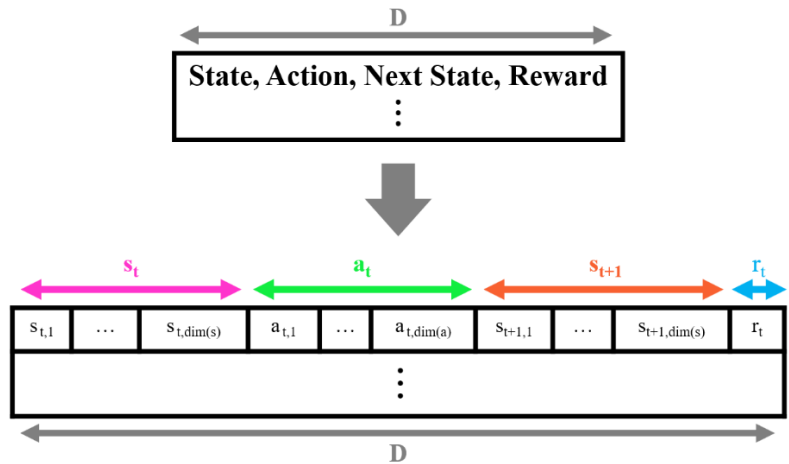

Figure 2: The placement of each transition quadruple in each row of the buffer

When the DRL members of each generation are being executed in the execution environment, the sequence of the experiences of each member – including the transition quadruple $(s_t, a_t, s_{t+1}, r_t)$ – are stored in their respective buffers. After the completion of all executions, a specific number of the best experience sequences of each member is selected according to their total obtained reward and are joined to create a three-dimensional auxiliary buffer named the "reference buffer." As can be seen in Figure 3, the reference buffer is a D*H*P buffer, where D indicates the length and dimension of the transition quadruple, H is the number of execution steps in a complete execution, and P is the number of selected experience sequences from the buffers of the members of the population. Each two-dimensional plane that creates the dimension P indicates one sequence of experiences from one complete execution of one member. The dimensions of the quadruple components of transition in the execution environments of this field of study are different. In other words, the dimensions for states and actions are different (and always more than one) and are determined based on the environment, while the reward is always one-dimensional. Therefore, the first dimension of the reference buffer, which is equal to the length of the transition quadruple, is, in fact, equal to two times the state dimension (i.e., current state and the next state) plus the action dimension plus one (i.e., the reward dimension). Following the creation of the reference buffer, each element of the buffer is replaced by the square of its difference from the mean value of its corresponding elements along the dimension P. Next, by averaging, the elements representing components of more than one dimension, such as state, action, and the next state in each transition quadruple, are mapped to one element in all planes of experience sequences. In other words, a new buffer with dimensions 4*H*P is created from the reference buffer by averaging and transforming all state-related columns to one column in all planes and repeating this process for action and next state columns.

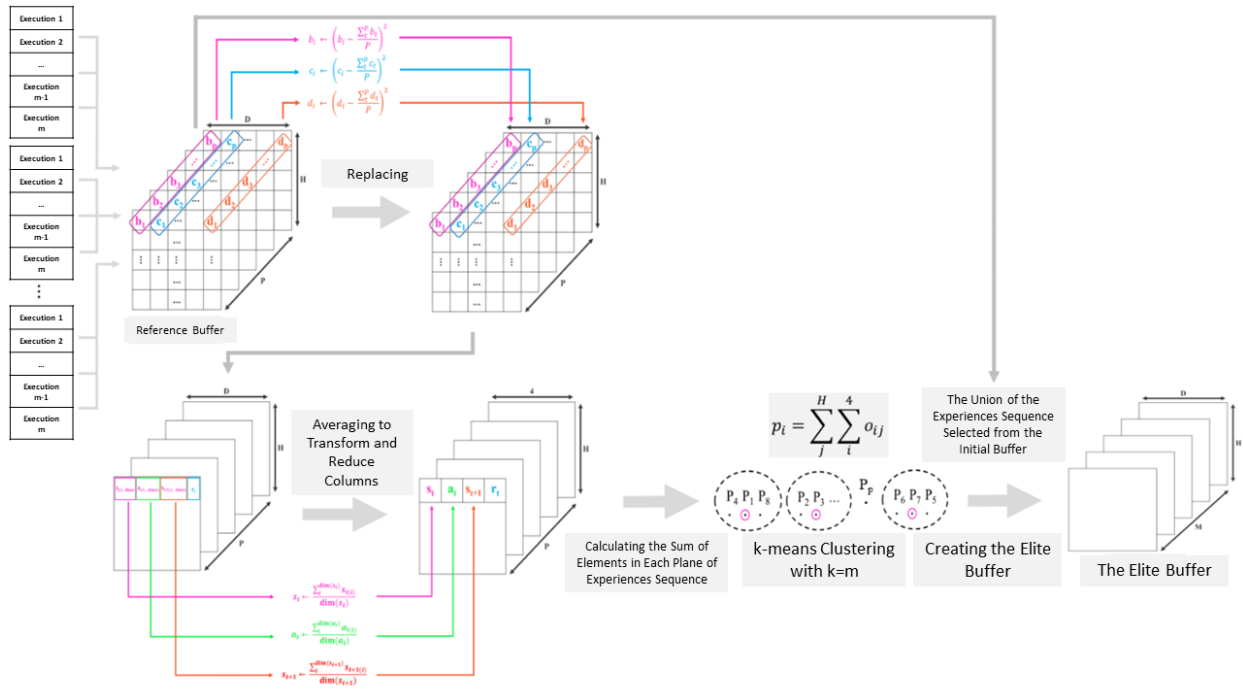

Figure 3: The mechanism and steps to create the elite buffer

Through the execution of the process described in the following, we attempt to select a specific number of the most different sequences of experiences from the existing sequences. Then, the sum of all elements in each plane of the sequence of experiences (i.e., each two-dimensional plane that constitutes the dimension P) is calculated. Each number resulting from the sum of the elements of each plane will represent that sequence of experiences. These representatives will be considered points in a one-dimensional space and clustered using the k-means clustering method. The value for k is equal to the number of complete executions of each member in each generation (M). Next, the nearest representative point to the center of the cluster (in each cluster) is introduced as one of the selected points. Finally, the sequences of the experiences corresponding to the points of the selected representatives are separated from the reference buffer and create a new D*H*M buffer entitled the "elite buffer." This buffer is used to improve the elite DRL network in each generation and to create hypothetical and interactive experiences for the mentioned network. The mechanism for the creation of the elite buffer and its constituting steps are shown in Figure3.

### 3.2.4. The Crossover Operator

In every generation, following the assessment of members, the establishment of the fitness function, and the selection of parents, the crossover operator is applied to each pair of parents. The aim is to execute the one-point crossover operator on the parameters of the actor network of each pair of parents with a predetermined probability. In the two born models, one of the other parameters is completely inherited from the first parent, and the other is completely inherited from the second parent. An illustrative example of the execution of the one-point crossover on the flattened parameters of the actor networks of each pair of parents is shown in Figure 4.

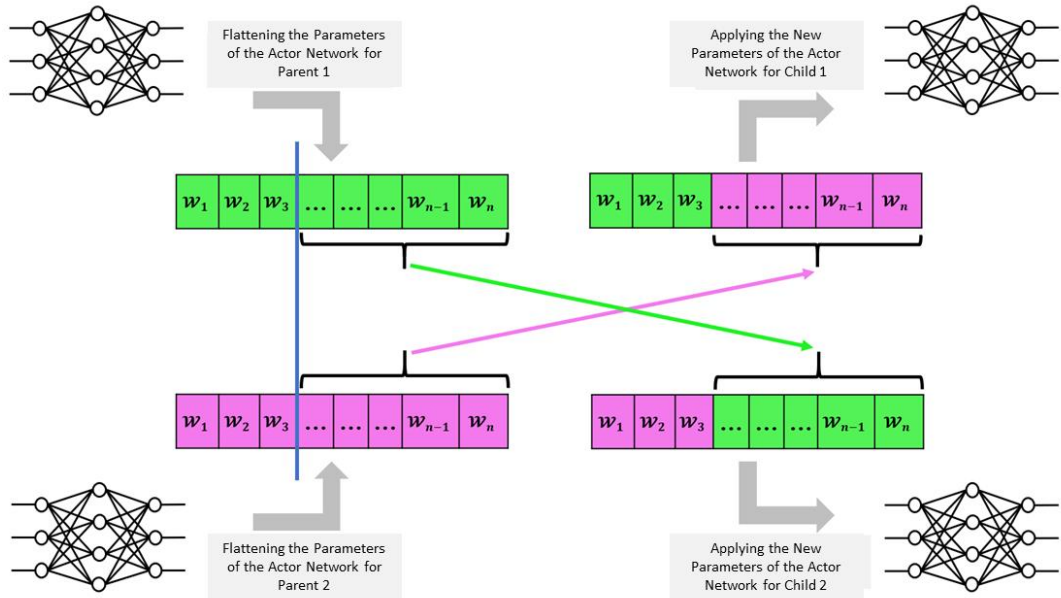

Figure 4: An example of executing the one-point crossover operator on the parameters of the actor networks of a pair of parents

### 3.2.5. The Mutation Operator

The mutation operator is executed on each born member with a probability. In other words, this operator is a Gaussian noise added to the born actor network parameters. To this aim, a random Gaussian noise, with the size of the parameters of the actor network, an average value of zero, and a standard deviation of one, is created for each member. Then, to avoid sudden fluctuations in the model, values more than +0.5 and less than -0.5 will be replaced by zero. Finally, the resulting noise is added to the born actor network parameters as the mutation operator. Figure 5 illustrates the process of creating the noise matrix (i.e., the mutation operator) in the form of a simple example [21].

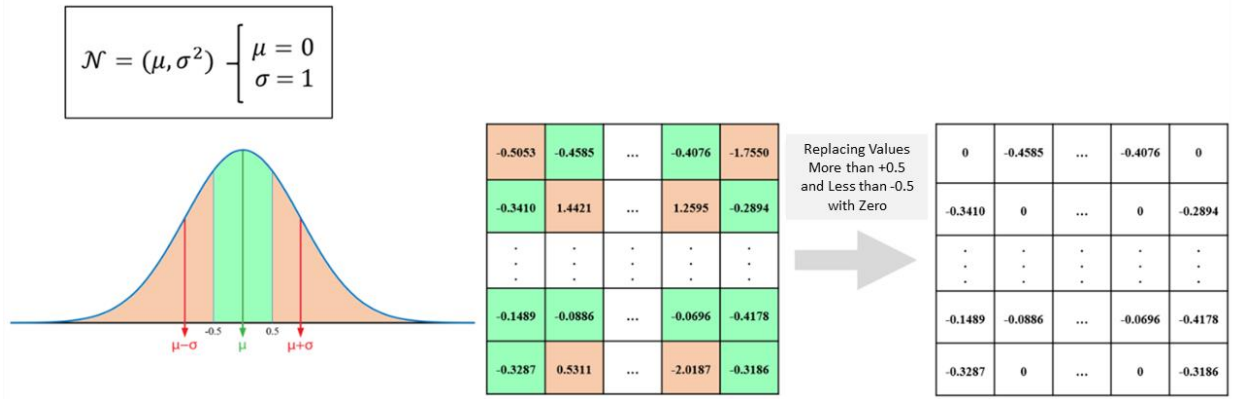

Figure 5: The process of generating noise matrix as the mutation operator

### 3.2.6. The Pseudocode

For further clarification, the pseudocode for the proposed method (EDRL-EB) is provided in Algorithm 1:

**Algorithm 1** EDRL-EB

Initialize a population of k DRL $pop$ with weight $\theta^\pi$, $\theta^{\pi'}$, $\theta^Q$ and $\theta^{Q'}$ respectively
Initialize replay buffers $R$
**for** $generation = 0, generation\_no$ **do**
    **for** $drl \in pop$ **do**
        **for** $episode = 0, episode\_no$ **do**
            Explore $environment$ using $drl$ via $Alg$ 2.1 or $Alg$ 2.2
            Append transition to corresponding replay buffer $R$ respectively
        **end for**
        $fitness$ = Evaluate $(drl, eval\_no)$
    **end for**
    Create $elite\ buffer$
    Select the $elite\ drl$ based on fitness score
    Create a copy of $elite\ drl$ and train it with $elite\ buffer$
    Select $(k-2)\ drls$ based on fitness score via roulette wheel selection
    Use $single - point\ crossover$ between two randomly selected parents on their $\theta^\pi$ from
        $(k-2)$ selected $drls$ under $crossover\_probability$ and next insert into next
        generation's population $pop'$
    **for** $(k-2)drl \in pop'$ **do**
        **if** $rand - no < mutation\_probability$ **then**
            add noise to $\theta^\pi$
        **end if**
    **end for**
    Insert the $elite\ drl$ into $pop'$
    Insert the copied $elite\ drl$ into $pop'$
    $pop \leftarrow pop'$
**end for**

As previously discussed, the EDRL-EB algorithm uses either DDPG or TD3 algorithms (which are called as Alg 2.1 and Alg 2.2 in Algorithm 1, respectively) for its DRL members. The pseudocodes for these two algorithms are provided in [18] and [19], respectively. Moreover, the pseudocode for the "Evaluate" function, which is called in the main part of the EDRL-EB algorithm, is shown in Algorithm 3.

---

**Algorithm 3** Function Evaluate

**procedure** Evaluate $(drl, eval\_no)$

    $fitness = 0$

    **for** $i = 0, eval\_no$ **do**

        Reset $environment$ and get initial state $s_0$

        **while** $env$ is not done **do**

            Select action $a_t = \pi(s_t, \theta^\pi) + noise_t$

            Execute action $a_t$ and observe reward $r_t$ and new state $s_{t+1}$

            $fitness \leftarrow fitness + r_t$ and $s = s_{t+1}$

        **end while**

    **end for**

    Return $\frac{fitness}{eval\_no}$

**end procedure**

---

## 4. Experiments and Results

### 4.1. Simulator

For the evaluation and assessment of the results of the proposed algorithm, we have employed the MuJuCo simulator. In addition to numerous key features and advantages that are beyond the scope of this paper, the models used in this simulator include elements that increase the complexity of the execution environment in continuous control problems and provide an appropriate platform to execute and assess the methods proposed in this field. To employ this simulator in the fields of reinforcement learning and deep reinforcement learning, the OpenAI Gym toolkit is required. Despite the existence of other simulators in this field, the superior features and capabilities of MuJoCo provided popularity and increased utilization of this simulator in the literature, and therefore, it was selected as the simulator for this study [23-27].

### 4.2. Execution Environments

Each environment possesses a Max Time Step parameter that indicates the maximum number of steps (i.e., actions) an actor carries out in the environment in each execution iteration. Based on the conditions of the environment and the executing actor, the actor possesses the capacity to consistently execute completely in some environments and cannot in some others. Based on the idea of constructing and utilizing the elite buffer in the proposed method, the elite buffer is created using the experience buffers of actors that have executed the Max Time Step and possess an equal length of buffer for each execution in the considered environment. Therefore, we need environments that can obtain the Max Time Step in each execution in order to execute and evaluate the proposed method. These environments are: HalfCheetah-v2, HumanoidStandup-v2, Swimmer-v2, and Reacher-v2, which all possess the appropriate complexity and adequate environmental diversity. Figure 6 illustrates a general view of these environments.

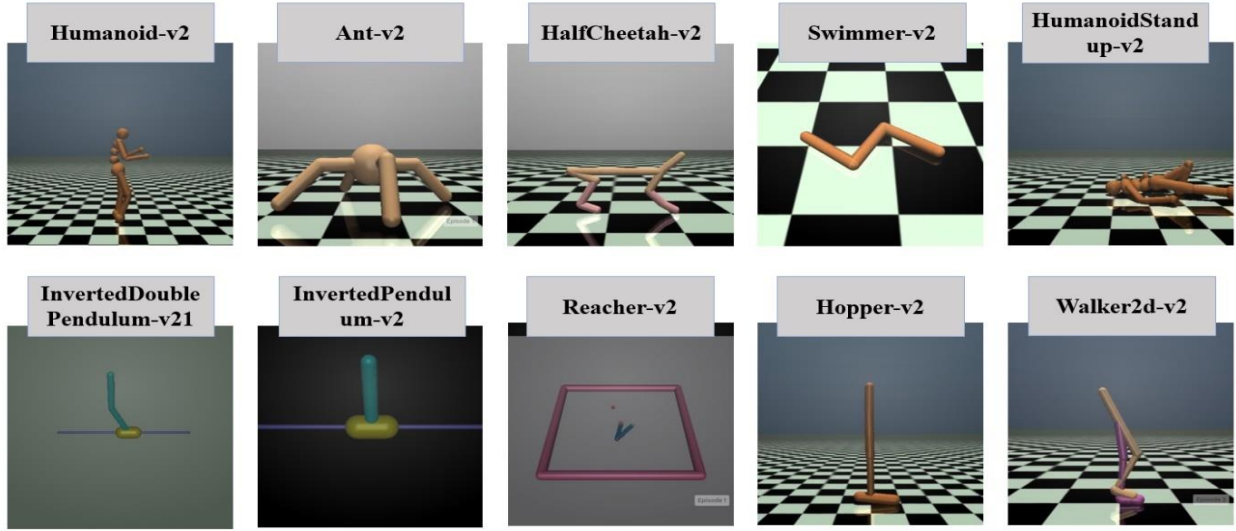

Figure 6: Continuous control environments in the MuJoCo Simulator – the snapshots of environments are taken from [28]

### 4.3. Assessment Criteria

After completing the execution of the EDRL-EB method, based on the nature of the problem, the total obtained reward for the elite member of the last generation is introduced as the performance score of the method. Similarly, in other methods, the total obtained reward of the model proposed as the solution at the end of the execution of the algorithm will be its performance score. However, due to the random nature of evolution-oriented algorithms and the effect of statistics and initial parameters on deep reinforcement learning models, the investigation of the results of a single execution of these methods will not provide a fair assessment. Therefore, for the evaluation of our considered methods, each was executed multiple (but equal) times, and its results are stored. Then, using statistical concepts such as average, variance, standard deviation, and median, results are analyzed.

In addition to assessment using statistical methods and standard criteria that are fitting for this field of research, the utilization of some qualitative and descriptive criteria, including the training flow and process, brittleness of the solution, and fluctuations, are possible in the assessment of the final results and diagrams, which will be provided in the following sections.

### 4.4. Results

The assessment of the proposed EDRL-EB method is carried out in two paths. First, the proposed method is investigated alone in accordance with the four main components. Then, after gaining a better view of how to yield higher success, the EDRL-EB method is evaluated and compared with some other notable methods in this field.

### 4.4.1. Assessing the Different States of the Proposed Method

In general, the EDRL-EB algorithm possesses four significant components, including the DRL model, elite buffer, crossover operator, and mutation operator. As mentioned earlier, the mechanism for this algorithm allows the exploitation of both DDPG and TD3 deep reinforcement learning methods. In other words, the members of the population in the algorithm can be one of these two models. On the other hand, the main component proposed in this method is the elite buffer and its manner of utilization. In addition, due to its evolutionary nature and the design of two operators (i.e., crossover and mutation) for this method, these two operators play key roles in the training method and the process of the algorithm.

In this section, we assess our proposed method by evaluating a number of general states of the mentioned components to yield an initial estimation of the success and effectiveness of these components in the final result.

In this assessment, nine different approaches to the EDRL-EB algorithm were executed in the test environments. Regardless of the nature of members, the EDRL-EB algorithm is capable of execution in three forms based on the main components: 1) using only the elite buffer without evolutionary operators (i.e., the probability of the execution of crossover and mutation is set to zero), 2) using the elite buffer with the mutation operator (i.e., the probability of the execution of crossover is set to zero), and 3) using the elite buffer with crossover and mutation. It should be noted that there is another execution form where the elite buffer is only used with the crossover operator (i.e., the probability of the execution of mutation is set to zero). However, in theory, due to the presence of sudden fluctuations caused by the chosen crossover operator and the inability to create the desired controlled diversity in the parameters of the DRL model, this form did not seem appropriate, and it was, therefore, declined as a candidate for further evaluation. In addition, execution in all three forms is possible using either DDPG or TD3. In this research, the parameters and architectures employed for DRL models are equivalent to the standard models introduced in their respective original papers. On the other hand, a new form of improved architecture for the DDPG network was studied and assessed in the paper that proposed TD3, which they claimed to have eliminated some of its problems and yielded better performance compared with the standard DDPG. Therefore, in addition to comparing the proposed method with the standard DDPG and TD3 deep reinforcement models, this section includes a comparison with the "DDPG Plus" architecture proposed in that study.

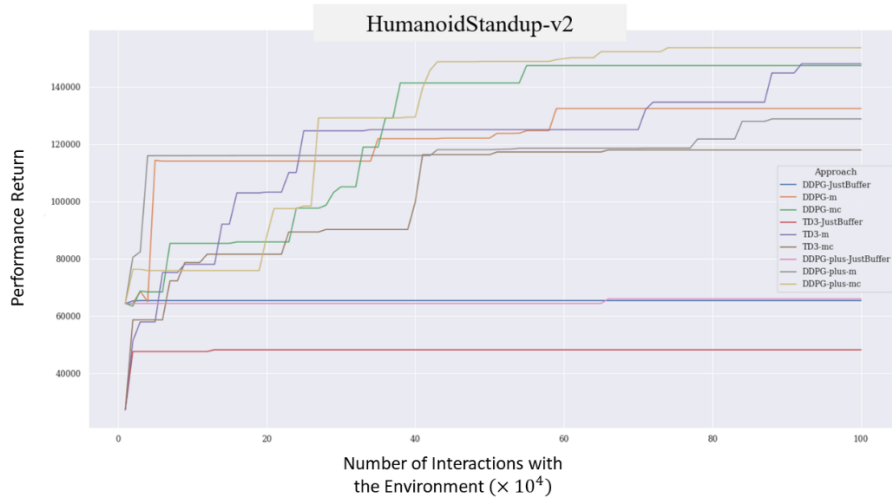

Figure 7: The results of assessing nine different approaches to the EDRL-EB algorithm in HumanoidStandup-v2

The results obtained for this assessment on two of the most challenging tests (i.e., HumaniodStandup-v2 and Swimmer-v2) are shown in Figures 7 and 8, respectively.

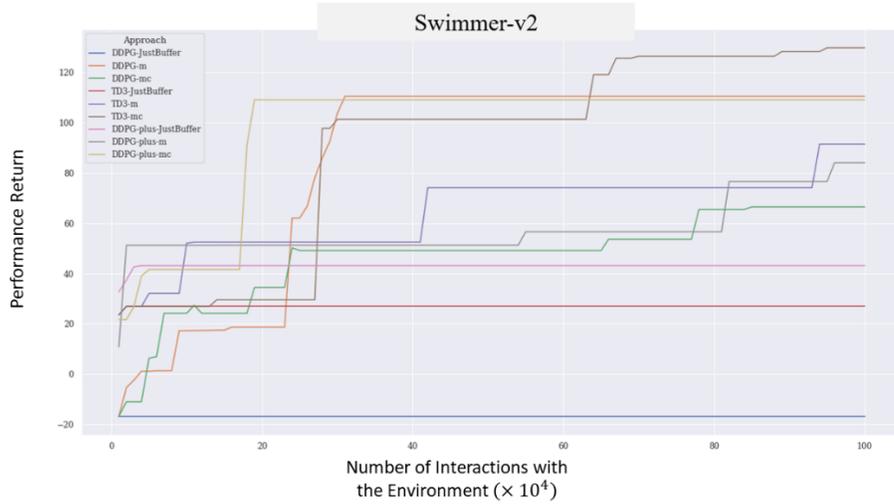

Figure 8: The results of assessing nine different approaches to the EDRL-EB algorithm in Swimmer-v2

For further clarification, the total obtained rewards for each of the nine mentioned approaches in the two environments are presented in Table 1. It should be noted that the final obtained reward of the method is determined after selecting the best member of the last generation (i.e., the "final elite member" or the "solution to the problem") and averaging the obtained rewards from multiple it's executions in the respective environment.

Table 1: The results obtained for the assessment of the proposed forms of the EDRL-EB algorithm in HumanoidStandup-v2 and Swimmer-v2 environments

| The Assessed Form | Respective Name in the Diagram | HumanoidStandup-v2 | Swimmer-v2 |
|---|---|---|---|
| DDPG with elite buffer | DDPG-JustBuffer | 65445.36 | -16.84 |
| DDPG with elite buffer and mutation | DDPG-m | 132461.01 | 110.51 |
| DDPG with elite buffer and crossover and mutation | DDPG-mc | 147462.36 | 66.51 |
| TD3 with elite buffer | TD3-JustBuffer | 48258.41 | 27.00 |
| TD3 with elite buffer and mutation | TD3-m | **148079.91** | 91.53 |
| TD3 with elite buffer and crossover and mutation | TD3-mc | 117991.44 | **129.87** |
| DDPG plus with elite buffer | DDPG-plus-JustBuffer | 66047.15 | 43.14 |
| DDPG plus with elite buffer and mutation | DDPG-plus-m | 128822.86 | 84.11 |
| DDPG plus with elite buffer and crossover and mutation | DDPG-plus-mc | **153654.81** | 109.17 |

As can be seen in the data from Table 1 and diagrams, in Swimmer-v2, the TD3-mc form (indicating the use of the TD3 model with the elite buffer and both operators) reported the highest total obtained reward (129.8671), and it was the most successful compared with other forms. Moreover, as can be seen in the diagram relative to the Swimmer-v2 environment, the training course in this state was ascending, and it continued the training until the end of the execution, except for some ranges in the middle of execution.

Moreover, in the HumanoidStandup-v2 environment, the DDPG-plus-mc form (indicating the use of the "DDPG Plus" model with the elite buffer and both operators) and TD3-m (indicating the use of the TD3 model with the elite buffer and the mutation operator) were first and second modes among the nine approaches, respectively, with total obtained rewards of 153654.8109 and 148079.9109, in respective order.

Even though the TD3-m mode obtained the second-highest total obtained reward in the HumanoidStandup-v2 environment, due to a number of reasons, including high final obtained reward, less interrupted (and continuous) ascending flow, and a more robust theory that resolved the instability due to overestimation bias in DDPG, the TD3 mode with the elite buffer and the application of the operators was used as the basis to assess the EDRL-EB algorithm and the base method to perform the evaluation with other algorithms.

The other notable issue in this experiment is that in all test environments, regardless of the employed DRL model, all three execution modes without operators were the last in terms of the total obtained reward. This occurrence emphasizes the positive and successful effect of the utilization of evolutionary methods in this field of research.

### 4.4.2. Assessment and Comparison of the Proposed Method with Relative Works

In this section, the proposed EDRL-EB method is compared with other methods, including DDPG, TD3, and CEM-RL. Among the previous works, multi-objective DRL methods and CERL (due to severe problems) and ERL (due to the significant superiority of CEM-RL compared to this method in all aspects) were excluded from the assessment list.

- **The HumanoidStandup-v2 Environment**

The HumanoidStandup-v2 environment is among the most complex environments in 3D continuous control simulators. The high number of state and action dimensions in this environment provides a very large search space for the actor's trial and error. Despite the existence of this complexity, as well as the significance and similarity of these environments to the real world, this method was not reported in most previous studies as a successful environment, which indicates the inadequate success of previous works.

The statistical results obtained in the HumanoidStandup-v2 environment are provided in Table 2. Moreover, the diagram for the performance return of the mentioned algorithms in the training process in this environment is shown in Figure 9.

Table 2: The results obtained for EDRL-EB, CEM-RL, TD3, and DDPG algorithms in the HumanoidStandup-v2 environment

| Algorithm | EDRL-EB | CEM-RL | TD3 | DDPG |
|---|---|---|---|---|
| Mean | **142016.25** | 29748.91 | 119489.81 | 104395.64 |
| Standard Deviation | 15515.65 | 9.47 | 41104.59 | 100.34 |
| Median | 142533.19 | 29749.49 | 147863.49 | 97178.31 |
| Best Result | 164812.46 | 29761.51 | 153990.11 | 117058.84 |

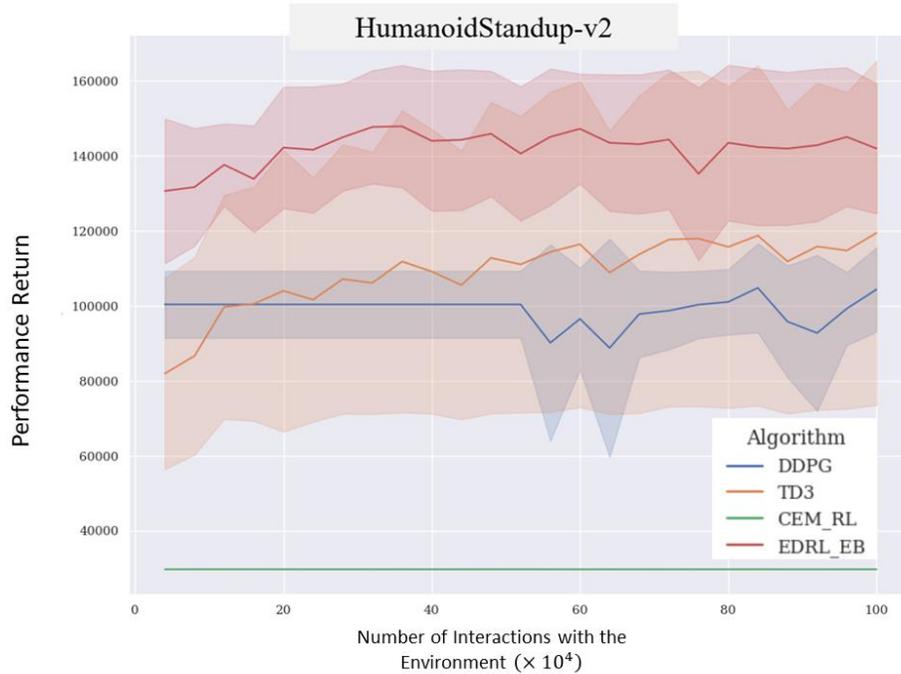

Figure 9: The diagram for the assessment of EDRL-EB, CEM-RL, TD3, and DDPG algorithms in the HumanoidStandup-v2 environment

As can be seen, the EDRL-EB method performs significantly better in the HumanoidStandup-v2 environment compared with other algorithms, followed by TD3 and DDPG in the second and third places (with a significant difference), respectively, and CEM-RL is the last. As is evident in the diagram, CEM-RL does not have any advancement during training in this environment and is not trained appropriately. The difference between the performance return of EDRL-EB and TD3 (i.e., 22526.44) is so significant that it clearly and vividly demonstrates the success of EDRL-EB in this environment. Moreover, other statistical indicators for this method confirm that the results for various executions have a desirable level of performance return. At the same time, appropriate standard deviation and median values show a high level of correlation and the absence of severe and consecutive fluctuations. It should be noted that even though the standard deviation in EDRL-EB is more than methods such as TD3, the significant difference of mean value in the performance return and the presence of design perspective in this study have rendered this issue insignificant. In addition, based on Figure 9, CEM-RL did not enter the training process altogether, DDPG started with an average total obtained reward but remained stationary and unchanged until almost halfway through the process flow and entered the training process (and its corresponding fluctuations) very late, and EDRL-EB and TD3 followed an ascending flow and were engaged in the training process. However, as mentioned earlier, there is a significant difference between the performance return of the two methods throughout the reported range.

- **The Swimmer-v2 Environment**

The Swimmer-v2 environment is one of the most well-known environments in the MuJoCo simulator. The relatively low dimensions of action and state in this environment and the existence of only two joints have made it look like a simple and straightforward environment. Despite the fact that Swimmer-v2 is indeed a very simple environment compared with the complex HumanoidStandup-v2 environment, its high sensitivity, rigorousness, and sparsity of the distribution of rewards have yielded the successful execution and the obtaining of a high performance return in this environment a very difficult task. In addition,

Swimmer-v2 is also a very challenging environment due to the presence of high fluctuations, brittle convergence, and randomness.

The statistical results obtained in the Swimmer-v2 environment are provided in Table 3. Moreover, the diagram for the performance return of the mentioned algorithms in the training process in this environment is shown in Figure 10.

Table 3: The results obtained for EDRL-EB, CEM-RL, TD3, and DDPG algorithms in the Swimmer-v2 environment

| Algorithm | EDRL-EB | CEM-RL | TD3 | DDPG |
|---|---|---|---|---|
| Mean | **248.56** | 242.6 | 97.81 | 31.53 |
| Standard Deviation | **74.59** | 106.06 | 13.65 | 9.24 |
| Median | 272.64 | 200.9 | 98 | 35.54 |
| Best Result | 329.14 | 366.56 | 115.56 | 38.83 |

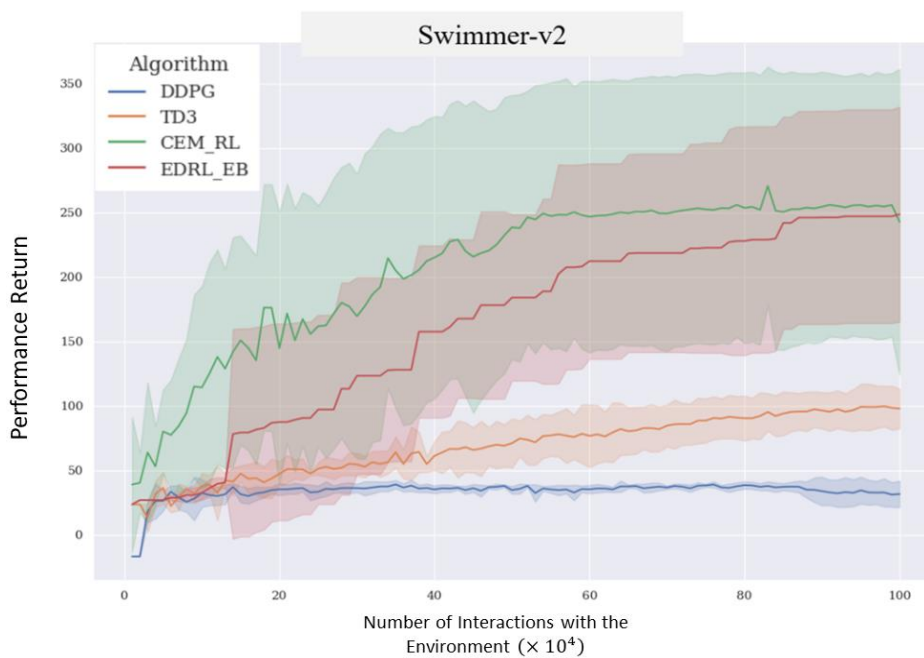

Figure 10: The diagram for the assessment of EDRL-EB, CEM-RL, TD3, and DDPG algorithms in the Swimmer-v2 environment

As can be seen in the results for the Swimmer-v2 environment, the EDRL-EB offers the best performance among the other methods with a small difference from CEM-RL. Both methods offer an appropriate mean performance return. However, the variance and standard deviation of EDRL-EB are significantly less than CEM-RL, indicating a higher sparsity and distance in the results for the executions of CEM-RL. In other words, CEM-RL functions mostly randomly in the Swimmer-v2 environment and possesses a higher dependence on the varying parameters in different executions, which increases the sparsity and distance between performance returns (determined using variance and standard deviation). In addition, the median is another advantageous statistical indicator for the performance analysis of these two methods. In EDRL-EB, the median is larger than the mean, while they are the opposite in CEM-RL (i.e., the median is less than the mean value). Median indicates that more than half of executions using EDRL-EB have performance returns more than the obtained mean value. Therefore, the peak for the distribution diagram of the executions of this method is more than the mean value and towards a higher performance return (a negative skewness). On the contrary, the results for the CEM-RL are the opposite.

On the other hand, as can be seen in the training process diagram, it is evident that EDRL-EB maintains its ascending trend throughout the execution and is continuously learning. As for the CEM-RL method, while it starts with a better performance return and maintains its significant superiority with EDRL-EB towards the mid-point of the trend, it loses its ascending learning trend later and halts around some specific performance returns. In addition to the halting of the learning process of the actor, this method shows a brittle convergence towards the end of the execution process, where it experiences a descending trend. Another notable issue is the learning speed of the methods. As is clearly seen in the learning process diagram, the velocity of the ascending trend in EDRL-EB is higher than its closest competitor (i.e., CEM-RL). Furthermore, the other two methods demonstrate a weak performance in this environment. To elaborate, TD3 has a very low velocity in its ascending learning trend and records a very low performance return at the end, and DDPG remains fixed around one point throughout the execution process and is not trained.

To summarize, in the Swimmer-v2 environment, considering all statistical indicators and qualitative indicators, including the reliability of the response, brittle convergence, skewness of the distribution, and trend and velocity of learning, the EDRL-EB method performs superior compared with other methods. Following this method, CEM-RL offers an appropriate and acceptable flow, while the other two methods perform very weak.

- **The HalfCheetah-v2 Environment**

The absence of complexities that are hard to comprehend, the presence of a proper visualization, and the ease of understanding the learning process are among the reasons that yield popularity for this environment. With 5 joints, 6 action dimensions, and 17 state dimensions, the HalfCheetah-v2 environment is among the continuous control environments that offer sufficient complexity and tangible performance. Moreover, due to the existence of a fixed body between the frontal part (i.e., the head and hands) and the back part (i.e., feet), which yields a fixed connection between the two moving parts, the action search space and the degree of freedom of the cheetah's joints are significantly limited compared with the real world. This limited movement has yielded increased ease and eliminated many local optima that were prone to trap the response.

The statistical results obtained in the HalfCheetah-v2 environment are provided in Table 4. Moreover, the diagram for the performance return of the mentioned algorithms in the training process in this environment is shown in Figure 11.

Table 4: The results obtained for EDRL-EB, CEM-RL, TD3, and DDPG algorithms in the HalfCheetah-v2 environment

| Algorithm | EDRL-EB | CEM-RL | TD3 | DDPG |
|---|---|---|---|---|
| **Mean** | 9911.97 | **11923.58** | 9777.45 | 8327.71 |
| **Standard Deviation** | **317.31** | 570.85 | 503.07 | 403.67 |
| **Median** | 10074.14 | 12084.98 | 9704.14 | 8246.34 |
| **Best Result** | 10225.6 | 12515.83 | 10519.98 | 8874.83 |

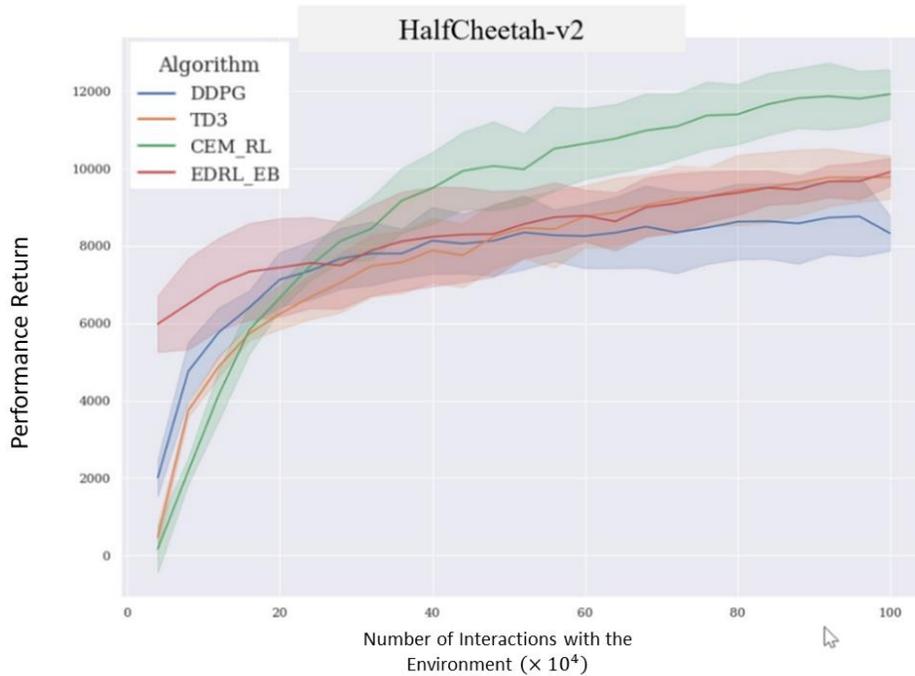

Figure 11: The diagram for the assessment of EDRL-EB, CEM-RL, TD3, and DDPG algorithms in the HalfCheetah-v2 environment

By obtaining the highest average performance return, the CEM-RL method is superior to all other assessed methods in the HalfCheetah-v2 environment. The high significant difference between the efficiency of this method and other methods indicates the appropriate success of CEM-RL in this environment. Following CEM-RL, the second, third, and fourth algorithms are EDRL-EB, TD3, and DDPG, respectively. Moreover, the fact that the median of the performance return of CEM-RL is more than the calculated mean value indicates the distribution with negative skewness and the presence of the distribution peak on the side of higher performance return.

Even though the EDRL-EB method obtains a desirable result, lower standard deviation compared with CEM-RL, negative skewness, the distribution function peak on the side of higher performance return, ascending learning trend, and higher initial performance return than other methods, it demonstrates inadequate velocity in the training and learning process and does not perform as expected by obtaining a final performance return less than CEM-RL. According to the conditions of the simulated cheetah in the HalfCheetah-v2 environment, the existence of very few local optimum points and the limited degree of freedom, as well as the slow ascending trend of the EDRL-EB method, two hypotheses can be expressed. The probability of improving the performance return of EDRL-EB in this environment by modifying its mechanism or increasing the probability of crossover and mutation operators (to provide better exploration and diversity in the training process) is the first hypothesis. On the other hand, the second hypothesis indicates that if the learning process continues with more steps and a higher number of interactions with the environment, it will eventually obtain superior performance return in an equal number of interactions with the environment due to some reasons such as the possession of an ascending learning trend, initiation of the process with a performance return higher than other methods, and absence of halt in the learning process.

While TD3 is marginally inferior to the EDRL-EB method in terms of performance return, it performs worse in other indicators, including higher standard deviation, positive skewness in the execution distribution, and decreasing learning velocity towards the end of the learning process. Finally, in this environment, the DDPG method obtains the last place. While this method initially has a high learning velocity and an adequate ascending trend, it loses its velocity very soon. From the mid-process to the end, this method experiences an increase in the range of the performance return and halts at a point, which hinders its capacity to learn. Finally, towards the end of the process, DDPG experiences brittleness.

- **The Reacher-v2 Environment**

The appearance and features of the actor in the Reacher-v2 environment are significantly simpler than in other environments. The actor in this environment has only one joint, two action dimensions, and 11 state dimensions. The external and objective simplicity of the environment and the high state-action dimension ratio (which offers an extremely larger state search space compared with its small action search space) yields a specific level of complexity for this environment. In addition to its specific complexity, the execution speed of transactions in this environment is very high. In other words, the complete execution of the environment includes only 50 time steps. The statistical results obtained in the Reacher-v2 environment are provided in Table 5. Moreover, the diagram for the performance return of the mentioned algorithms in the training process in this environment is shown in Figure 12.

Table 5: The results obtained for EDRL-EB, CEM-RL, TD3, and DDPG algorithms in the Reacher-v2 environment

| Algorithm | EDRL-EB | CEM-RL | TD3 | DDPG |
|---|---|---|---|---|
| Mean | -5.42 | **-3.99** | -5.03 | -10.2 |
| Standard Deviation | 0.12 | 1.6 | 0.2 | 0.84 |
| Median | -5.36 | -2.99 | -5.05 | -9.96 |
| Best Result | -5.34 | -1.04 | -4.96 | -9.21 |

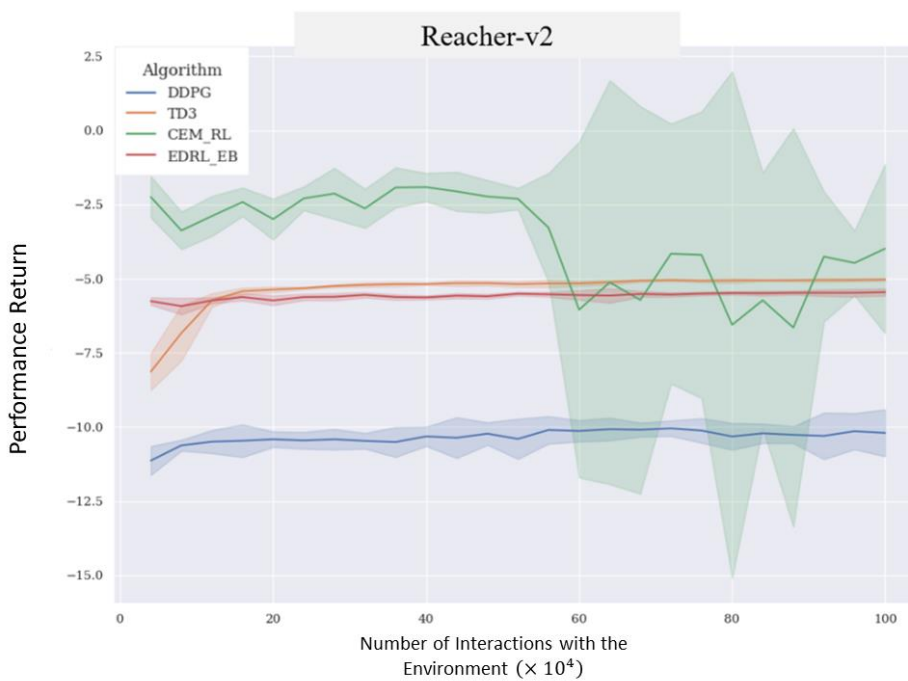

Figure 12: The diagram for the assessment of EDRL-EB, CEM-RL, TD3, and DDPG algorithms in the Reacher-v2 environment

As can be seen in the diagram and statistical indicators, the CEM-RL method obtains a higher average performance return in the Reacher-v2 environment. In this environment, the performance return starts at a negative value, and it needs to increase during training and learning to obtain a higher reward. Before assessing the statistical indicators of this environment, it should be noted that the learning process diagrams in this environment clearly show that the learning process of none of the assessed algorithms shows a natural and improving process. In other words, none of the assessed methods are able to learn and maximize their performance return in the Reacher-v2 environment, and they all halt at specific points. Even though CEM-RL yields a superior final performance return, other assessment indicators, including variance, standard deviation, and other intuitive and qualitative indicators shown in the learning process diagram, show the unsuccessful execution of this method in this environment. To elaborate, it is evident that CEM-RL starts with a higher performance return but experiences continuous and severe fluctuations despite not having a constant learning trend. Moreover, as it passes the mid-learning process, fluctuations and standard deviation for different executions increase significantly. These fluctuations, along with the reduction in the average performance return, show a clear brittle convergence of CEM-RL in this environment. On the other hand, even though EDRL-EB records a lower average performance return and better standard deviation, it does not perform successfully in the Reacher-v2 environment (similar to its competitors) since it does not learn in this environment.

In general, none of the methods in the fields of DRL, ERL, and EDRL perform successfully in the Reacher-v2 environment, and they all encounter problems such as halting at a fixed point and no effective interactions with the training process. However, this environment is among the continuous control environments in the MuJoCo simulator, and it possesses similarities to this field. Moreover, this environment has differences in structure and dimensions, and it has a very short execution time and a maximum time step. Therefore, since all the mentioned methods failed globally in this environment, it seems that studying other methods in the field of reinforcement learning can improve, or at least propose, ideas and approaches to work with this environment and environments similar to this. The methods assessed in this study are methods with complex calculations that require a large amount of data, function well in complex environments, and have a long learning process. Hence, it seems that the Reacher-v2 environment does not inherently require such complex methods and cannot provide the necessary high-dimensional data for these methods. Consequently, studying and assessing methods with less complexity that require fewer necessities in the field of reinforcement learning can be beneficial in obtaining solutions to yield improvement in the Reacher-v2 environment.

## 5. Discussion

In the previous section, we evaluated the results obtained for each algorithm in separate test environments and provided their outcomes in tables and learning process diagrams. Then, we thoroughly discussed the performance of each algorithm in terms of statistical indicators such as mean, variance, standard deviation, median, distribution skewness, and best performance, along with qualitative indicators such as the brittleness of convergence, self-destructiveness, learning trend and velocity, fluctuations during learning, and halting during learning. In addition, the results for success or failure of the EDRL-EB method were discussed in detail, along with further discussions on reasons or existing hypotheses. In "Table A" provided in Appendix 1, the summary and notable issues in the mentioned assessments are provided.

As discussed earlier, the DDPG algorithm generally faces problems and deficiencies, especially in overestimation bias. These problems were evident in the failure of this method to obtain success in the above tests and the emergence of problems such as incorrect learning, weak performance return, brittle convergence, and halting at some point.

The TD3 algorithm is a very powerful method that has shown proper capability and performance in many environments and problems. However, in addition to the problems it has in the general field of professional continuous control, it generally suffers from facing environments and problems with high levels of complexity and dimensions, failing to carry out effective exploration in problems with high complexity, relying on the initial parameters of the problems, and managing execution fluctuations.

The performance of the CEM-RL algorithm has been bipolar in our assessments. In environments and problems that do not possess features such as a very high number of dimensions and complexities (e.g., the HumanoidStandup-v2 environment), numerous local optima, and a high degree of freedom in joints, this algorithm completes the training process with an ascending trend and a reasonable velocity and obtains a proper performance return. On the other hand, its performance is very weak in problems with a very high number of dimensions and complexities, numerous local optima, a high degree of freedom in joints, sparse reward, or problems that require facing risks for effective exploration in the environment. In such conditions, this algorithm cannot offer proper learning. In specific, in such environments and problems (which share the most similarity with real-world problems), the major problems and deficiencies of this algorithm include a complete stop or halt, minimal performance return, severe fluctuations, brittle convergence, and self-destruction.

Based on the provided results, if the final performance return is strictly considered, the performance of the proposed EDRL-EB method is very similar to that of the CEM-RL. However, by performing a thorough investigation and considering other indicators and conditions of the problem, we realize that EDRL-EB is significantly superior. By decreasing dependency and sensitivity to the initial parameters of the reinforcement learning components, utilizing interactive knowledge and learning while employing the learning and assessment of hypothetical experiences, and carrying out effective exploration using proper evolutionary operators, the EDRL-EB method possesses a very desirable capability and performance in problems with high complexities and very high data dimensions (e.g., the HumanoidStandup-v2 environment). This capability is evident by comparing its superior performance return in that environment with other competitors. Other capabilities of this method include: learning with a proper velocity and with an ascending trend, fewer fluctuations in the learning process, and not halting learning in the middle of the process and – most times – not even at the end of the process. To see a link to videos for the execution of the EDRL-EB algorithm, please refer to Appendix 2.

In addition to the successful performance of the proposed method in terms of statistical and qualitative indicators mentioned earlier, a more careful investigation reveals that EDRL-EB performed very well in eliminating the general challenges in the field of deep reinforcement learning mentioned at the beginning of this paper. We should mention that challenges such as temporal credit assignment with sparse reward, absence of effective exploration, brittle convergence, and insufficient performance return in problems with high complexity and dimensions are among the most significant challenges in the field of deep reinforcement learning and – especially – continuous control. As shown in the results and their relative discussions, the success of the EDRL-EB method in offering an outstanding improvement in robust convergence, reduction of fluctuations, and improvement of performance return in problems with high levels of complexities and dimensions is evident. On the other hand, a significant improvement in the performance return, a significant decrease in complications with local optima and unproductive fluctuations, its execution process, and the accuracy of the final model in the test environments indicate better and more effective exploration in the investigated environments. Therefore, it is evident that EDRL-EB successfully obtained better and more effective exploration using the elite buffer and collective wisdom (inspired by the interactive learning mechanism and hypothetical experiences of the human brain) in conjunction with evolutionary operators. Furthermore, as discussed earlier, temporal credit assignment and eligibility trace in the human brain are carried out using a complex and ambiguous – but very accurate – mechanism. In the EDRL-EB method, without directly considering a precise mechanism to resolve this problem, the possibility of indirect training, and consequently, reusing experiences that include actions with very high rewards with delays are provided using the elite buffer (which includes the best and most different sequences of experiences of the population), without accurately identifying their temporal relationships. Using this method, we can exploit their advantages in improving the policy of the selected model without the need to accurately identify these actions or understand their type and delayed performance manner.

It should be noted that due to the significant influence of hyperparameters in the execution process and the performance return of the algorithm, parameter tuning is crucial. If this part of modeling is not carried out accurately and properly, we may not obtain desirable results even when using a method that possesses a good power in terms of theory and proposed mechanism. In other words, obtaining and manifesting the true power of these methods require the accurate completion of the parameter tuning step. In the present paper, this step was not fulfilled appropriately due to the existing limitations, and it still requires further work. Therefore, based on the reasons, documents, the theory of the proposed method, and the results of the assessments, the possibility of improving the performance of the EDRL-EB method and obtaining better success is certain through a better tuning of parameters. It should be noted that while this algorithm requires further work to tune its parameters, a more investigative study towards maturing the mechanism of using hypothetical and selected experiences in the elite buffer can yield better and more extensive performance and capabilities.

## 6. Conclusion

The EDRL-EB algorithm is a method proposed to alleviate challenges and problems in the field of continuous control deep reinforcement learning. This study attempted to combine deep reinforcement learning methods with evolutionary computations, and in addition to presenting an idea inspired by the process of interactive learning and hypothetical experiences in the human brain, aimed to improve the performance return in the field of continuous control.

According to the results and discussed assessment indicators, the EDRL-EB method performed superbly towards alleviating the general challenges in the field of deep reinforcement learning discussed at the beginning of the present paper. Moreover, due to the crucial significance of hyperparameters in the execution process and the performance return of the algorithm, further investigation and experiments towards obtaining ideal hyperparameters and maturing the elite buffer and evolutionary operators employed in this study can significantly improve the manifestations of the power and efficiency of the algorithm. In addition, these improvements provide the possibility for a more extensive competition with the state-of-the-art methods in this field of research. Hence, the future direction of this study will be investigating and maturing the algorithm.

# Appendix 1

Table A provides an integrated and summarized discussion of the note-worthy issues in the assessments provided in Section 4. In this table, the boxes related to the issues of each method in the test environment are colored based on the degree of success and the overall performance of the algorithms. In specific, green, cyan, blue, amber, coral, and pink indicate the best to worst possible performance in the mentioned environment in respective order.

Table A: A summary of the assessment of EDRL-EB, CEM-RL, TD3, and DDPG algorithms in the test environments

| Environment | DDPG | TD3 | CEM-RL | EDRL-EB |
|---|---|---|---|---|
| HumanoidStandup-v2 | • Weak performance return | • Moderate downward performance return<br>• Distribution with negative skewness and peak more than mean | • Very weak performance return | • Very high performance return<br>• Distribution with negative skewness and peak more than mean |
| | • Prolonged stillness and a very long delay at the beginning of the occurrence of learning<br>• High fluctuations and lack of convergence<br>• Absence of the occurrence of learning | • Occurrence of learning<br>• Ascending trend with low velocity and presence of fluctuations<br>• Non-stop learning until the end of the process | • Absence of the occurrence of learning<br>• Complete stillness with approximately minimal performance return | • Ascending trend with good velocity<br>• Occurrence of learning<br>• Possibility to improve by tuning parameters<br>• Non-stop learning until the end of the process |
| Swimmer-v2 | • Very weak performance return | • Weak performance return | • High performance return<br>• High standard deviation<br>• Distribution with positive skewness and peak less than mean | • High performance return<br>• Low standard deviation<br>• Distribution with negative skewness and peak more than mean |
| | • Absence of the occurrence of learning<br>• Stillness with a very weak performance return | • Ascending trend with a very low velocity | • Halting the learning process from mid-point<br>• Brittle convergence | • Non-stop learning until the end of the process<br>• Occurrence of learning<br>• Ascending learning trend with high velocity and no fluctuations |
| HalfCheetah-v2 | • Weak performance return<br>• Distribution with positive skewness and peak less than mean | • Moderate performance return<br>• Standard deviation more than EDRL-EB | • High performance return<br>• Standard deviation more than EDRL-EB | • Moderate performance return<br>• Low standard deviation<br>• Distribution with negative skewness |

|  |  | • Distribution with near-zero skewness |  | and peak more than mean |
| --- | --- | --- | --- | --- |
|  | • Halting the learning process from mid-point<br>• Brittle convergence | • Occurrence of learning<br>• Non-stop learning until the end of the process<br>• Ascending learning trend with moderate velocity | • Occurrence of learning<br>• Non-stop learning until the end of the process<br>• Ascending learning trend with good velocity and no fluctuations | • Occurrence of learning<br>• Non-stop learning until the end of the process<br>• Ascending learning trend with moderate velocity<br>• Possibility to improve by tuning parameters |
| Reacher-v2 | • Very weak performance return | • Weak performance return | • Weak performance return<br>• High standard deviation | • Weak performance return |
|  | • Absence of the occurrence of learning<br>• Stillness with a very weak performance return | • Absence of the occurrence of learning<br>• Stillness with a very weak performance return | • Absence of the occurrence of learning<br>• Brittle convergence<br>• High fluctuations<br>• Self-destruction | • Absence of the occurrence of learning<br>• Stillness with a very weak performance return |

## Appendix 2

The video for the execution of the EDRL-EB algorithm in the four test environments can be found in the following link:

https://github.com/mesma1995/EDRL-EB